\documentclass[10pt,twocolumn,letterpaper]{article}

\usepackage{iccv}
\usepackage{times}
\usepackage{epsfig}
\usepackage{graphicx}
\usepackage{amsmath}
\usepackage{amssymb}
\usepackage{colortbl}
\usepackage{cuted}
\usepackage{capt-of}
\usepackage[linesnumbered, boxed, ruled]{algorithm2e}

\usepackage{graphicx}
\usepackage{amsmath}
\usepackage{bm}
\usepackage{amssymb}
\usepackage{booktabs}
\usepackage[etex=true,export]{adjustbox}
\usepackage{graphicx, amsmath, amssymb, caption, subcaption, multirow, overpic, textpos}
\usepackage[table]{xcolor}
\usepackage{physics}
\usepackage{float}

\definecolor{citecolor}{HTML}{0071BC}
\definecolor{linkcolor}{HTML}{ED1C24}
\usepackage[pagebackref=true,breaklinks=true,letterpaper=true,colorlinks,citecolor=citecolor,linkcolor=linkcolor,bookmarks=false]{hyperref}

\iccvfinalcopy 

\def\iccvPaperID{3175} 

\ificcvfinal\fi

\definecolor{citecolor}{HTML}{0071BC}
\definecolor{linkcolor}{HTML}{ED1C24}
\definecolor{my_blue}{HTML}{0050F0}
\definecolor{my_purple}{HTML}{7030A0}
\definecolor{my_light}{HTML}{CAB2D6}
\definecolor{my_green}{HTML}{00C050}

\renewcommand{\paragraph}[1]{\vspace{1.25mm}\noindent\textbf{#1}}

\newcolumntype{x}[1]{>{\centering\arraybackslash}p{#1pt}}
\newcolumntype{y}[1]{>{\raggedright\arraybackslash}p{#1pt}}
\newcolumntype{z}[1]{>{\raggedleft\arraybackslash}p{#1pt}}

\newcommand{\app}{\raise.17ex\hbox{$\scriptstyle\sim$}}

\definecolor{deemph}{gray}{0.6}

\definecolor{baselinecolor}{gray}{.9}

\definecolor{my_red}{HTML}{FE4444}
\definecolor{Highlight}{HTML}{39b54a}  
\definecolor{GreenCell}{HTML}{e6fff7} 
\definecolor{DarkGray}{HTML}{777777} 
\definecolor{Gray}{gray}{0.95}

\begin{document}

\title{Video-P2P: Video Editing with Cross-attention Control}

\author{First Author\\
Institution1\\
Institution1 address\\
{\tt\small firstauthor@i1.org}
\and
Second Author\\
Institution2\\
First line of institution2 address\\
{\tt\small secondauthor@i2.org}
}
\author{Shaoteng Liu$^{1}$\hspace{1.0cm}Yuechen Zhang$^{1}$\hspace{1.0cm}Wenbo Li$^{1}$\hspace{1.0cm}Zhe Lin$^{3}$\hspace{1.0cm}Jiaya Jia$^{1,2}$\\
$^{1}$The Chinese University of Hong Kong~~~
$^{2}$SmartMore~~~
$^{3}$Adobe
\\
\url{https://video-p2p.github.io/}
}
\maketitle

\begin{strip}\centering
\vspace{-40pt}
\includegraphics[width=1\textwidth]{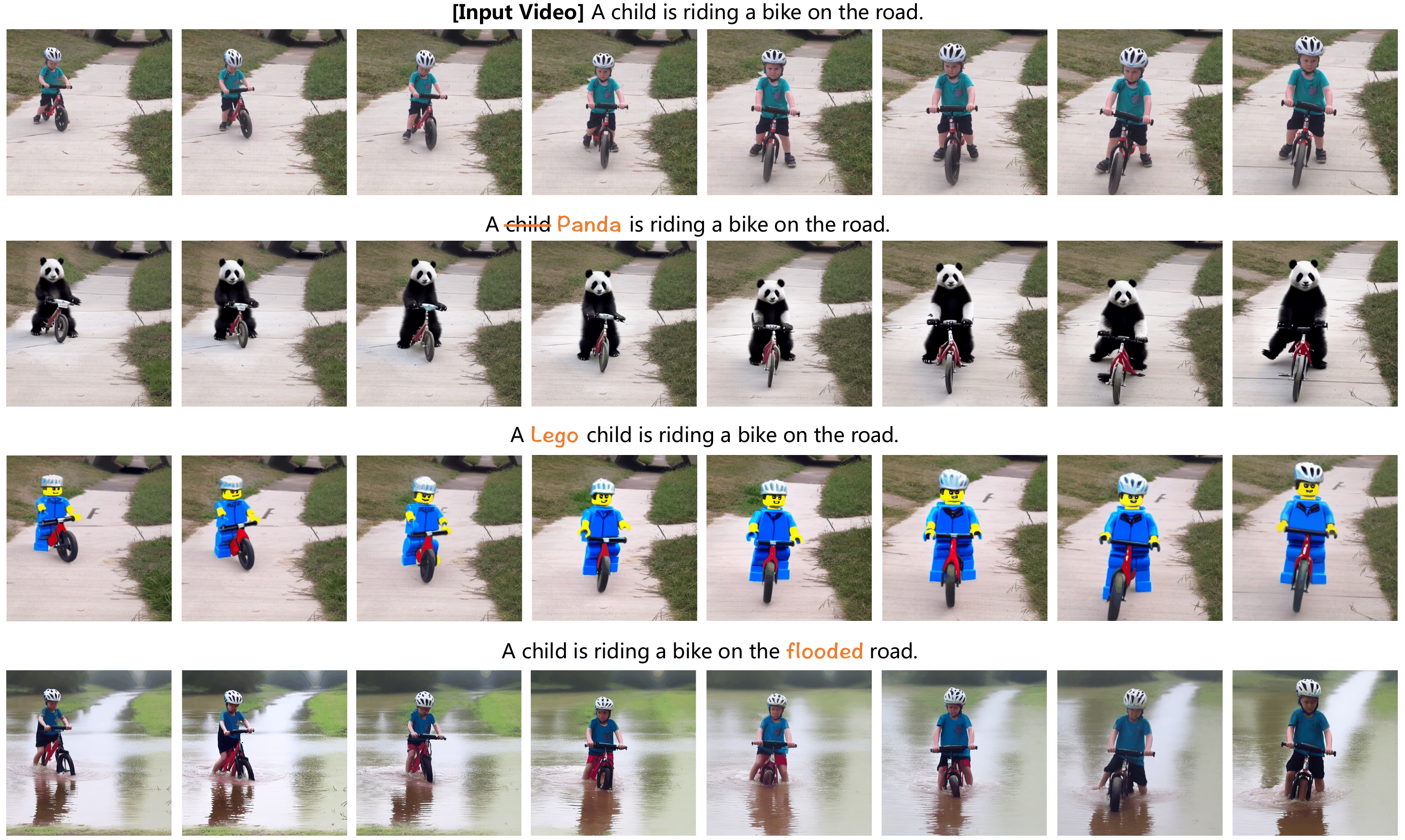}
\vspace{-5mm}
\captionof{figure}{
    Video-P2P generates new characters while optimally maintaining the pose and environment in videos.}
    \label{fig:teaser}
\end{strip}

\ificcvfinal\thispagestyle{plain}\fi

\begin{abstract}
\vspace{-2mm}
This paper presents Video-P2P, a novel framework for real-world video editing with cross-attention control. While attention control has proven effective for image editing with pre-trained image generation models, there are currently no large-scale video generation models publicly available. Video-P2P addresses this limitation by adapting an image generation diffusion model to complete various video editing tasks. Specifically, we propose to first tune a Text-to-Set (T2S) model to complete an approximate inversion and then optimize a shared unconditional embedding to achieve accurate video inversion with a small memory cost. For attention control, we introduce a novel decoupled-guidance strategy, which uses different guidance strategies for the source and target prompts. The optimized unconditional embedding for the source prompt improves reconstruction ability, while an initialized unconditional embedding for the target prompt enhances editability. Incorporating the attention maps of these two branches enables detailed editing. These technical designs enable various text-driven editing applications, including word swap, prompt refinement, and attention re-weighting. Video-P2P works well on real-world videos for generating new characters while optimally preserving their original poses and scenes. 
It significantly outperforms previous approaches.
\end{abstract}
\vspace{-5mm}
\section{Introduction}

Video creation and editing are key tasks~\cite{ho2022video,ho2022imagen,villegas2022phenaki,singer2022make,molad2023dreamix}. Text-driven editing becomes one promising pipeline. Several methods have demonstrated the ability to edit generated or real-world images with text prompts~\cite{kawar2022imagic,hertz2022prompt,mokady2022null}. Till now, it is still challenging to edit only local objects in a video, such as changing a running ``dog'' into a ``cat'' without influencing the environment. This paper proposes a pipeline that can edit a video both locally and globally, as shown in Figs.~\ref{fig:teaser} and \ref{fig:examples}.

Text-driven image editing requires a model capable of generating target content, such as changing the category or property of an object. Diffusion models have demonstrated outstanding generation capabilities in this area~\cite{kawar2022imagic,hertz2022prompt,brooks2022instructpix2pix,tumanyan2022plug}. Among these methods, attention control emerges as the most effective pipeline for detailed image editing~\cite{hertz2022prompt,mokady2022null}.
In order to edit a real image, this pipeline includes two necessary steps: (1) inverting the image into latent features with a pre-trained diffusion model, and (2) controlling attention maps in the denoising process to edit the corresponding parts of the image. For example, by swapping their attention maps, we can replace a ``child'' with a ``panda''.

In this paper, we aim to build an attention control-based pipeline for video editing. Since no large-scale pre-trained video generation models are publicly available, we propose a novel framework to show that a pre-trained image diffusion model can be adapted for detailed video editing.

\begin{figure}
\begin{center}
   \includegraphics[width=1\linewidth]{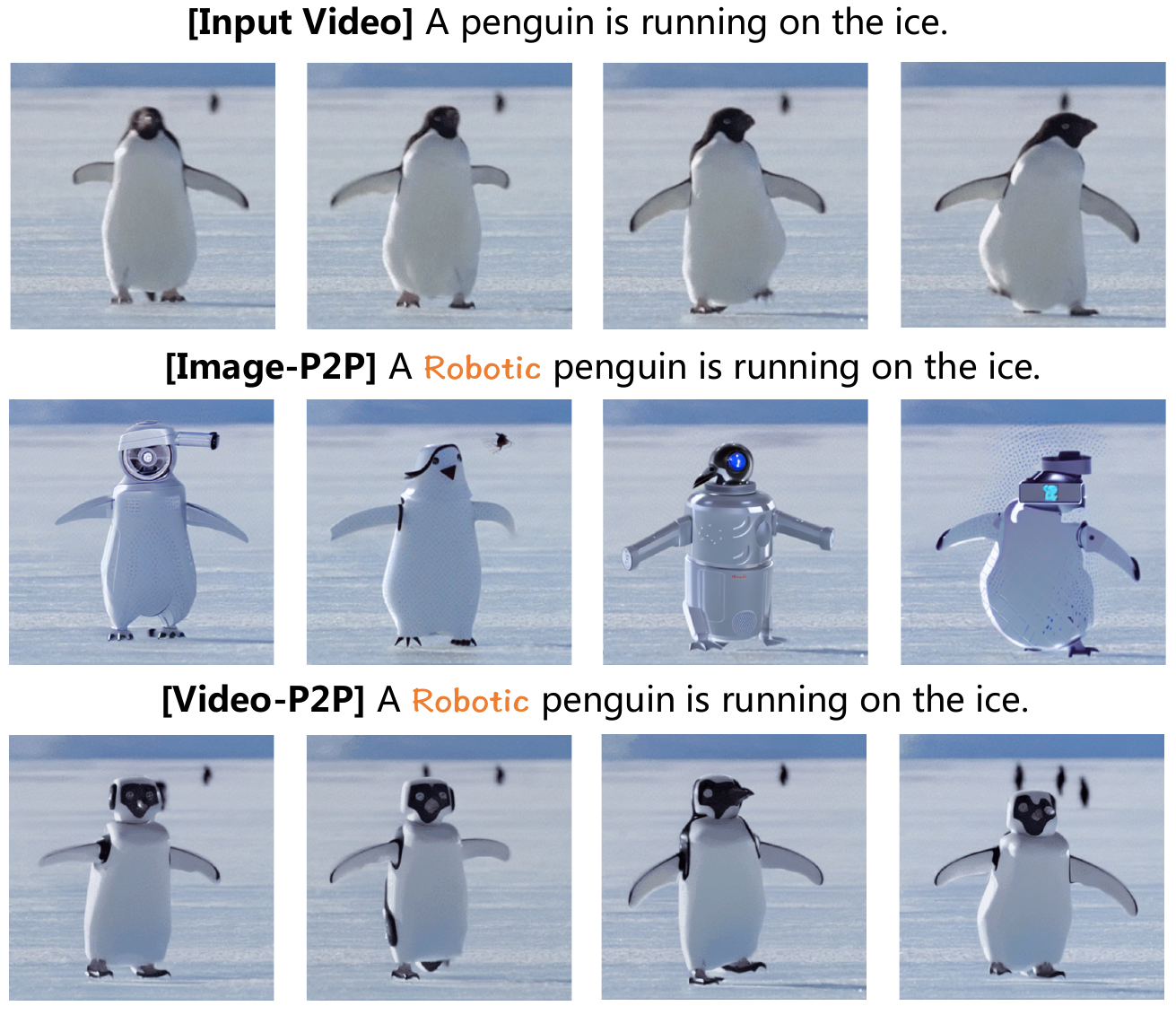}
\end{center}
   \vspace{-10pt}
   \caption{Video-P2P vs Image-P2P. Editing a video frame-by-frame (Image-P2P) cannot guarantee semantic consistency across frames. Video-P2P enables changing the penguin into the same robotic type in every frame.}
   \vspace{-10pt}
\label{fig:video_vs_image}
\end{figure}

While a pre-trained image diffusion model can be utilized for video editing by processing frames individually (Image-P2P), it lacks semantic consistency across frames (the 2nd row of Fig.~\ref{fig:video_vs_image}).
To maintain semantic consistency, we propose using a structure on inversion and attention control for all frames, by transforming the Text-to-image diffusion model (T2I) into a Text-to-set model (T2S).
This approach is effective, as illustrated in the 3rd row, where the robotic penguin maintains its consistency across frames.

\begin{figure*}[t]
\begin{center}
   \includegraphics[width=1\linewidth]{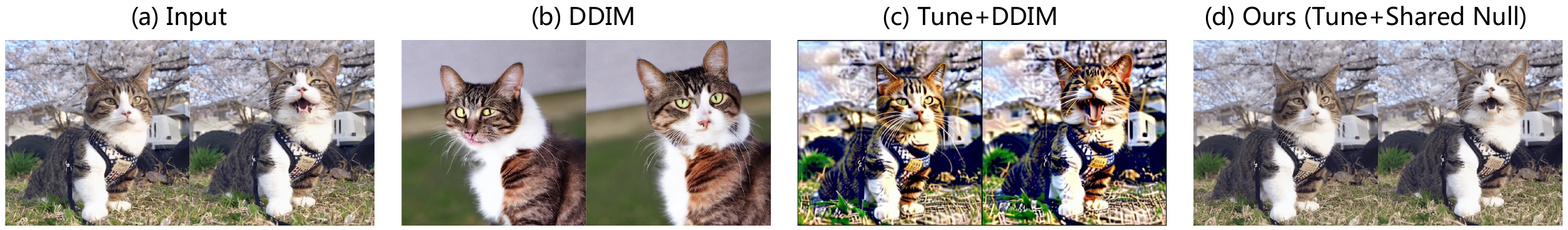}
\end{center}
   \vspace{-10pt}
   \caption{Inversion Comparison. (b) The inflated model cannot generate high-quality results. (c) Tuning the model can create an approximate video inversion. (d) Optimizing a shared unconditional embedding can accurately reconstruct the input video.}
   \vspace{-10pt}
\label{fig:inversion}
\end{figure*}

We adopt the method proposed in \cite{wu2022tune} to convert a T2I model into a T2S model by altering the convolution kernels and replacing the self-attentions with frame-attentions.
This conversion yields a model that generates a set of semantically consistent images.
The generation quality will be degraded with the inflation step but it can be recovered after tuning on the original video.
Although the tuned T2S model is not an ideal video generation model, it suffices to create an approximate inversion for a video as shown in Fig.~\ref{fig:inversion} (c). It is just an approximation because errors are accumulated in the denoising pass, consistent with conclusions in~\cite{mokady2022null,wallace2022edict}.

To improve the inversion quality, we propose to optimize a shared unconditional embedding for all frames to align the denoising latent features with the diffusion latent features.
Our experiments show that shared embedding is the most efficient and effective choice for video inversion. Comparisons are shown in Fig.~\ref{fig:inversion}.

As discussed in~\cite{hertz2022prompt}, successful attention control requires a model to have both reconstruction ability and editability.
While image inversion has been argued to possess both abilities in \cite{mokady2022null}, we find that video editing presents different challenges.
The T2S model, as an inflation model not trained on any videos, is not robust to the perturbations caused by various unconditional embeddings.
Although our optimized embedding can achieve reconstruction, changing prompts can destabilize the model and result in a low-quality generation.
On the other hand, we find that the approximate inversion with an initialized unconditional embedding is editable but cannot reconstruct well.
To address this issue, we propose a decoupled-guidance strategy in attention control, utilizing different guidance strategies for the source and target prompts. 
Specifically, we use the optimized unconditional embedding for the source prompt and the initialized unconditional embedding for the target prompt. We incorporate the attention maps from these two branches to generate the target video. These two simple designs prove effective and successfully complete video editing. Our contributions can be summarized as:

\begin{itemize}
\item {We propose the first framework for video editing with attention control. A decoupled-guidance strategy is designed to further improve performance.}
\item {We introduce an efficient and effective video inversion method with shared unconditional embedding optimization to improve video editing substantially.}
\item {We conduct extensive ablation studies and comparisons to show the effectiveness of our video editing framework.}
\end{itemize}

\section{Related Work}

\subsection{Text Driven Generation}
DALL-E~\cite{ramesh2021zero} first considers the text-to-image (T2I) generation task as a sequence-to-sequence translation problem, with subsequent research improving generation quality~\cite{yu2022scaling,ding2022cogview2,gafni2022make}. Denoising Diffusion Probabilistic Models (DDPMs)\cite{ho2020denoising} have recently gained popularity for T2I. GLIDE\cite{nichol2021glide} utilizes classifier-free guidance to improve text conditioning. DALLE-2~\cite{ramesh2022hierarchical} leverages CLIP~\cite{radford2021learning} for better text-image alignment. Latent Diffusion Models (LDMs)~\cite{rombach2022high} propose processing in the latent space to enhance training efficiency. In our work, we employ a pre-trained image diffusion model based on LDMs.

Text-to-video (T2V) generation is a nascent research area. GODIVA~\cite{wu2021godiva} first introduces VQ-VAE~\cite{van2017neural} to T2V. CogVideo~\cite{hong2022cogvideo} combines T2V with CogView-2~\cite{ding2022cogview2}, utilizing pre-trained text-to-image models. Video Diffusion Models (VDM)~\cite{ho2022video} propose a space-time U-Net for performing diffusion on pixels. Imagen Video~\cite{ho2022imagen} successfully generates high-quality videos with cascaded diffusion models and v-prediction parameterization.
Phenaki~\cite{villegas2022phenaki} generates videos with time-variable prompts.
Make-A-Video~\cite{singer2022make} combines the appearance generation of T2I models with movement information from video data.
While these approaches generate reasonable short videos, they still contain artifacts and do not support real-world video editing. Additionally, most of these approaches are not publicly available at this time.

Several single-video generative models have been proposed. Single-video GANs~\cite{arora2021singan,gur2020hierarchical} can generate novel videos with similar objects and motions, while SinFusion~\cite{nikankin2022sinfusion} uses diffusion models to improve generalization but is limited to simple cases. Tune-A-Video~\cite{wu2022tune} inflates an image diffusion model into a video model and tunes it to reconstruct the input video. It allows for changes in semantic content but with limited temporal consistency. We find that using DDIM inversion results can improve its temporal consistency. However, it cannot avoid altering unrelated regions.
We adapt some designs of TAV to do our model initialization.

\subsection{Text Driven Editing}
Generative models have demonstrated impressive performance in image editing, with approaches ranging from GANs~\cite{gal2022stylegan,park2019semantic,patashnik2021styleclip,wang2018high} to diffusion models~\cite{avrahami2022blended,kawar2022imagic}. SDEdit~\cite{meng2021sdedit} adds noise to an input image and uses the diffusion process to recover an edited version. Prompt-to-Prompt~\cite{hertz2022prompt} and Plug-and-Play~\cite{tumanyan2022plug} use attention control to minimize changes to unrelated parts, while Null-Text Inversion~\cite{mokady2022null} improves real image editing. InstructPix2Pix~\cite{brooks2022instructpix2pix} enables flexible text-driven editing with user-provided instructions. Textual Inversion~\cite{gal2022image}, DreamBooth~\cite{ruiz2022dreambooth}, and Custom-Diffusion~\cite{kumari2022multi} learn special tokens for personalized concepts and generate related images.

Video editing with generative models has seen several advances recently. Text2Live~\cite{bar2022text2live} employs CLIP to edit textures in videos but struggles with significant semantic changes. Dreamix~\cite{molad2023dreamix} uses a pre-trained Imagen Video~\cite{ho2022imagen} backbone to perform image-to-video and video-to-video editing, with the ability to change motion as well. Gen-1~\cite{esser2023structure} trains models jointly on images and videos for tasks such as stylization and customization.
While these methods enable modifying video content, they operate like guided generation and tend to modify all regions together when editing an object. Our proposed method allows for local editing with a diffusion model pre-trained on images.

\section{Method}

Let $\mathcal{V}$ be a real video containing $n$ frames. We adopt the Prompt-to-Prompt setting by introducing a source prompt $\mathcal{P}$ and an edited prompt $\mathcal{P}^*$ which together generate an edited video $\mathcal{V}^*$ containing $n$ frames. The prompts are provided by the user.

\begin{figure}
\begin{center}
   \includegraphics[width=1\linewidth]{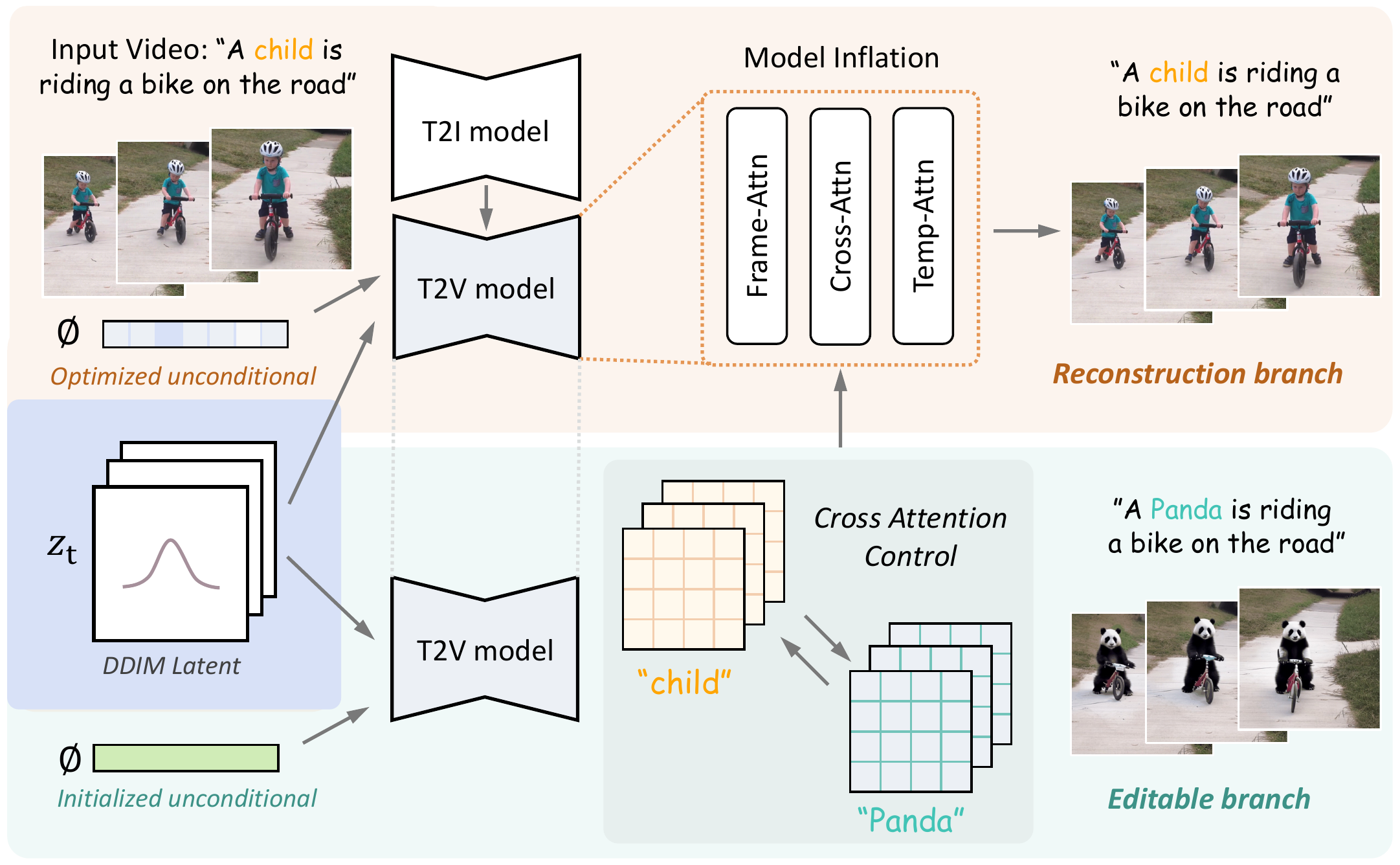}
\end{center}
\vspace{-10pt}
   \caption{Framework. We optimize one shared unconditional embedding for the reconstruct branch (orange). The initialized unconditional embedding is utilized for the editable branch (green). Their attention maps are incorporated to create the target video.}
\label{fig:framework}
\vspace{-10pt}
\end{figure}

To achieve cross-attention control in video editing, we propose Video-P2P, a framework with two key technical designs: (1) optimizing a shared unconditional embedding for video inversion, and (2) using different guidance for the source and edited prompts, and incorporating their attention maps. The framework is illustrated in Fig.~\ref{fig:framework}.

\subsection{Preliminary}
\paragraph{Latent Diffusion Models (LDMs).}
LDMs generate an image latent $z_0$ using a random noise vector $z_t$ and a textual condition $P$ as inputs.
As variants of DDPMs, these models aim to predict artificial noise by minimizing the following objective:
\begin{equation}
\min _\theta E_{z_0, \varepsilon \sim N(0, I), t \sim \text { Uniform }(1, T)}\left\|\varepsilon-\varepsilon_\theta\left(z_t, t, \mathcal{C}\right)\right\|_2^2,
\end{equation}
where $\mathcal{C}=\psi(\mathcal{P})$ is the embedding of the text prompt, and noise $\varepsilon$ is added to $z_0$ according to step $t$ to obtain $z_t$. During inference, the model predicts noise $\varepsilon_\theta(\cdot)$ for $T$ steps to generate an image from $z_T$.

\paragraph{DDIM sampling and inversion.}
Deterministic DDIM sampling can be used to generate an image from latent features in a small number of denoising steps:
\begin{equation}
\small
z_{t-1}=\sqrt{\frac{\alpha_{t-1}}{\alpha_t}} z_t+\left(\sqrt{\frac{1}{\alpha_{t-1}}-1}-\sqrt{\frac{1}{\alpha_t}-1}\right) \cdot \varepsilon_\theta\left(z_t, t, \mathcal{C}\right) .
\end{equation}
We use an encoder to encode the real image before the diffusion process and a decoder to decode after the denoising process. DDIM sampling can be reversed in a few steps through the equation:
\begin{equation}
\small
z_{t+1}=\sqrt{\frac{\alpha_{t+1}}{\alpha_t}} z_t+\left(\sqrt{\frac{1}{\alpha_{t+1}}-1}-\sqrt{\frac{1}{\alpha_t}-1}\right) \cdot \varepsilon_\theta\left(z_t, t, \mathcal{C}\right) ,
\end{equation}
known as DDIM inversion~\cite{song2020denoising}.
This can be used to obtain the corresponding latent features of a real image.

\paragraph{Null-text inversion.}
To mitigate the amplification effect of text conditioning during image generation, classifier-free guidance is proposed, which performs unconditional prediction~\cite {ho2022classifier}:
\begin{equation}
\tilde{\varepsilon}_\theta\left(z_t, t, \mathcal{C}, \varnothing\right)=w \cdot \varepsilon_\theta\left(z_t, t, \mathcal{C}\right)+(1-w) \cdot \varepsilon_\theta\left(z_t, t, \varnothing\right),
\end{equation}
where $\varnothing=\psi(" ")$ is the embedding of a null text and $w$ is the guidance weight.
However, the classifier-free guidance increases errors accumulated in the denoising process, leading to imperfect image reconstruction using the DDIM inversion.
\cite{mokady2022null} proposes to align the diffusion latent trajectory $z_T^*, \ldots, z_0^*$ with the denoising latent trajectory $z_T, \ldots, z_0$ by optimizing a step-wise unconditional embedding $\varnothing_t$:
\begin{equation}
\min _{\varnothing_t}\left\|z_{t-1}^*-z_{t-1}\right\|_2^2.
\end{equation}

\subsection{Video Inversion}
We begin by constructing a T2S model that is capable of performing an approximate inversion.
Following the VDM baselines~\cite{ho2022imagen,ho2022video} and TAV~\cite{wu2022tune}, we employ $1 \times 3 \times 3$ pattern convolution kernels and temporal attention. Moreover, we replace the self-attentions with frame-attentions, which take the first frames $v_0$ and the current frame $v_i$ as inputs and update features for the frame $v_i$. The formulation of the frame-attention is as follows:
\begin{equation}
Q=W^Q \mathbf{v}_i, K=W^K \mathbf{v}_0, V=W^V \mathbf{v}_0,
\end{equation}
where $W$ are the projection matrices in attention.
The model processes a video pair-by-pair and computes $n$ times to obtain the prediction for every frame.
While the Sparse-causal attention proposed in TAV~\cite{wu2022tune} outperforms frame-attention when generating videos from random noise, we find that the simple design suffices for video inversion since the reversed latent features can capture temporal information. Additionally, frame-attention conserves memory and speeds up the process.

While model inflation can aid in preserving semantic consistency across frames, it adversely impacts the generation quality of the T2I model. This is because the self-attention parameters are utilized to compute frame correlations, which have not been pre-trained.
Consequently, the T2S model, generated through inflation, is insufficient for the approximate inversion, as demonstrated in Fig.~\ref{tab:inversion_compare}.
To address this, we fine-tune the query projection matrices $W^Q$ of the frame- and cross-attentions, as well as additional temporal attention, to perform noise prediction based on the input video following~\cite{wu2022tune}.
After this initialization, the T2S model is capable of generating semantically consistent image sets while maintaining the quality of each frame, resulting in successful approximate inversion.

Using the fine-tuned T2S model, we perform video inversion by optimizing a shared unconditional embedding.
During inversion, each latent feature $z_t$ contains a channel for the frames with dimension $n$, where $z_{t,i}$ denotes the latent feature for the $i$-th frame. We employ DDIM inversion to generate latent features $z_0^*, \ldots, z_T^*$. The unconditional embedding is defined as follows:
\begin{equation}
\min _{\varnothing_t}\sum_{i=1}^{n}{\left\|z_{t-1, i}^*-z_{t-1, i}\left(\bar{z}_{t, i}, \bar{z}_{t, 0}, \varnothing_t, \mathcal{C}\right)\right\|_2^2} \mathrm{, where}
\end{equation}
\begin{equation}
\bar{z}_{t-1, i} = z_{t-1, i}\left(\bar{z}_{t, i}, \bar{z}_{t, 0}, \varnothing_t, \mathcal{C}\right)
\end{equation}
is updated at each step.
The T2S model's frame-attentions use two latent features to calculate the corresponding feature for the next step.
Notice $\varnothing_t$ is shared by all frames ($i=1,\ldots,n$) which minimizes the memory usage. Besides, using the same unconditional embedding for all frames avoids destabilizing the semantic consistency in attention control.

\subsection{Decoupled-guidance Attention Control}
To perform attention control on real images, existing works~\cite{hertz2022prompt,mokady2022null} require an inference pipeline with both reconstruction ability and editability.
However, achieving such a pipeline for a T2S model is challenging.
Video inversion allows us to establish an inference pipeline to reconstruct the original video well. However, the T2S model is not as robust as T2I models due to a lack of pre-training with videos. As a result, its editability is compromised with the optimized unconditional embedding, leading to degraded generation quality when changing prompts.
In contrast, we find that using an initialized unconditional embedding makes the model more editable while it cannot reconstruct perfectly.
This inspires us to combine the abilities of two inference pipelines.
For the source prompt, we use the optimized unconditional embedding in the classifier-free guidance. For the target prompt, we choose the initialized unconditional embedding.
We then incorporate attention maps from these two branches to obtain the edited video, where the unchanged parts are influenced by the source branch and the edited parts are influenced by the target branch.

\setlength{\textfloatsep}{7pt}
\begin{algorithm}[h]
\SetAlgoLined
\textbf{Input:} A source prompt $\mathcal{P}$, a target prompt $\mathcal{P}^*$.\\
\textbf{Output:} Source video $V_{src}$ and edited video $V_{dst}$.\\
 Latent features from DDIM inversion: $z_{T}$;\\
 $z_{T}^* \gets z_{T}$; \\
 Initialized unconditional embedding $\varnothing^*$ and optimized unconditional embedding $\varnothing$;\\
 \For{$t=T,T-1,\ldots,1$}{
    $z_{t-1}, M_{t} \gets DM(z_{t},\mathcal{P},t, \varnothing)$\;
    $M_{t}^* \gets DM(z_{t}^*,\mathcal{P}^*,t, \varnothing^*)$\;
    $\widehat{M}_{t} \gets Edit(M_{t}, M_{t}^*, t)$\;
    $z_{t-1}^* \gets DM(z_{t}^*,\mathcal{P}^*,t,\varnothing^*)\{M_{t}^* \gets \widehat{M}_{t}\}$\;
    $\alpha \leftarrow B\big(\overline{M}_{t, w}\big) \cup B\big(\overline{M}_{t, w^*}^*\big)$\;
    $z_{t-1}^* \leftarrow(1-\alpha) \odot z_{t-1}+\alpha \odot z_{t-1}^*$\;
 }
 \textbf{Return} $(z_{0},z_{0}^*)$
 \caption{Prompt-to-Prompt video editing}
 \label{alg:attention}
\end{algorithm}

The pseudo algorithm is shown in Alg.~\ref{alg:attention}.
We adopt the attention control methods from Image-P2P to Video-P2P.
For example, to perform word swap, the $Edit$ function can be represented as:
\begin{equation}
Edit\left(M_t, M_t^*, t\right):= \begin{cases}M_t^* & \text { if } t<\tau \\ M_t & \text { otherwise }\end{cases},
\end{equation}
$M_t$ and $M_t^*$ are the cross-attention maps for every frame at every step, and $DM$ is the tuned T2S model. Changing the frame-attentions maps has a small influence on the final results.
Attention maps are swapped only for the first $\tau$ steps because attentions are formed in the early period.
$\overline{M}_{t, w}$ is the average attention map of the word $w$ calculated at step $t$.
It is averaged over steps $T,\ldots,t$ independently for every frame. For the $j$-th frame, we calculate:
\begin{equation}
\overline{M}_{t, w, j} = \frac{1}{T-t} \sum_{i=t}^{T}{\overline{M}_{i, w, j}} \quad j = 1,\ldots,n.
\end{equation}
$B\left(\overline{M}_{t, w}\right)$ represents the binary mask obtained from the attention map. A value is set to 1 when larger than a threshold.


\begin{figure*}[htb]
\begin{center}
    \includegraphics[width=1\linewidth]{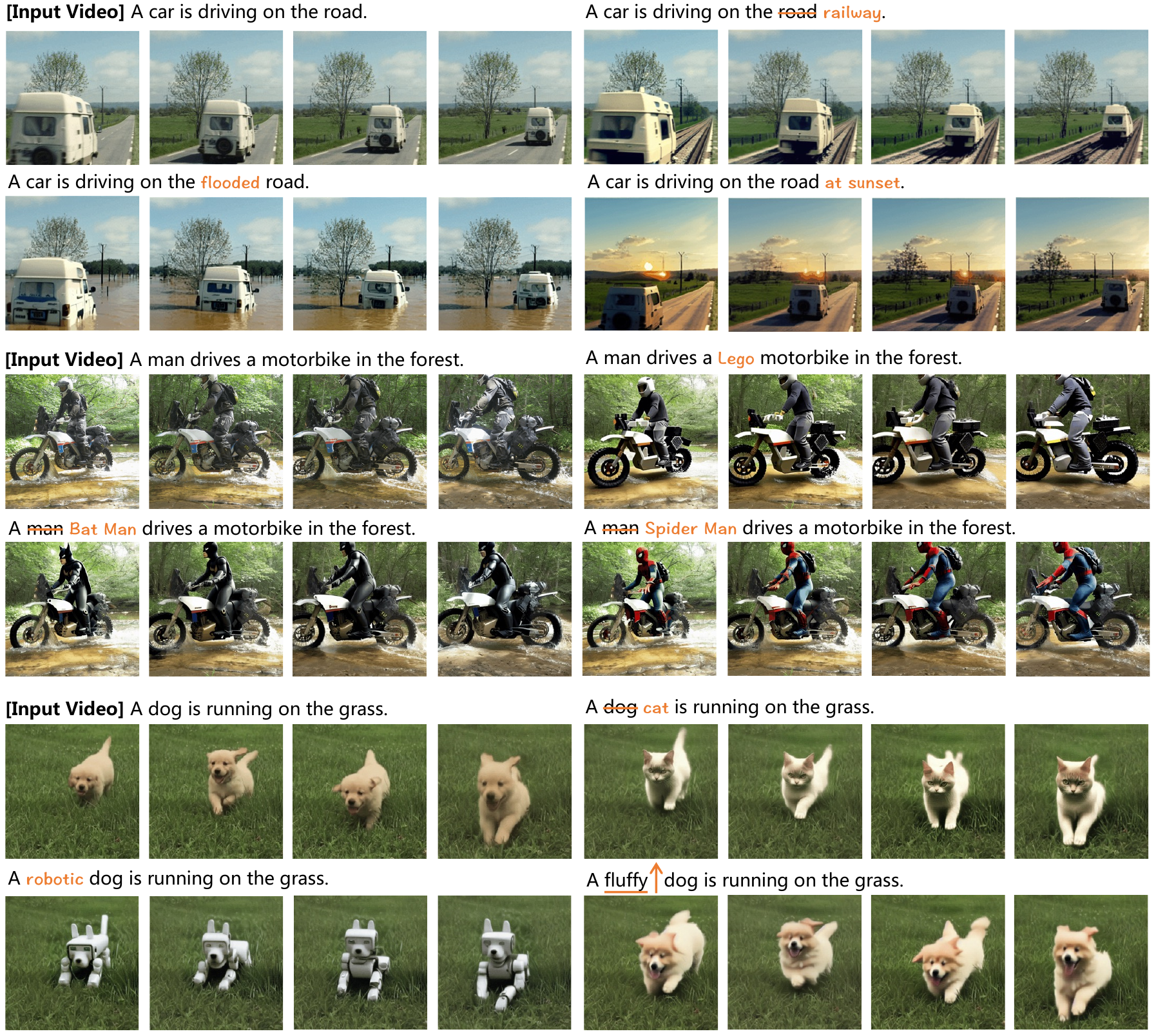}
\end{center}
\vspace{-13pt}
\caption{Videos edited by Video-P2P with text prompts. Video-P2P can do both word swaps and prompt refinement.}
\vspace{-10pt}
\label{fig:examples}
\end{figure*}

\begin{figure*}[t]
\begin{center}
    \includegraphics[width=1\linewidth]{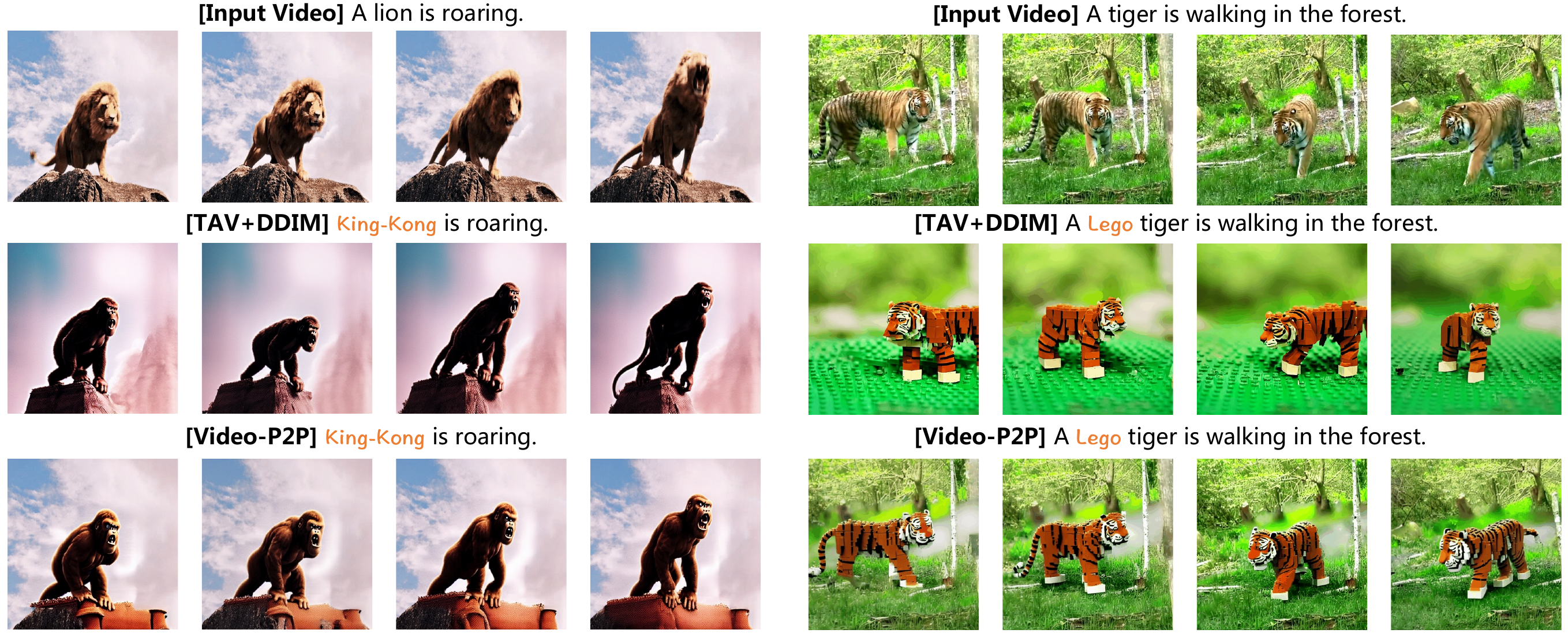}
\end{center}
\vspace{-12pt}
\caption{Video-P2P v.s. Tune-A-Video (TAV). Video-P2P offers the ability to edit content locally, while TAV+DDIM cannot avoid influencing unrelated regions.}
\label{fig:p2p_vs_tav}
\vspace{-12pt}
\end{figure*}

\section{Experiments}
\subsection{Implementation Details}
We develop our method based on CompVis Stable Diffusion (v1-5).
Similar to TAV~\cite{wu2022tune}, we fix the image autoencoder and sample 8 or 24 frames at the resolution of 512 $\times$ 512 from a video.
To initialize the model, we fine-tune the T2S model for 500 steps to reconstruct the original video.
During attention control, we set the cross-attention replacing ratio to 0.4 and the attention threshold to 0.3. For prompt refinement, we set the refinement ratio to 0.4. These parameters can be adjusted to control the editing fidelity for different examples.
All 8-frame experiments are conducted on a single V100 GPU, with 5 minutes for initialization (tuning), 6 minutes for inversion, and 1 minute for inference.

\subsection{Applications}
Our Video-P2P method can be utilized for a range of editing applications, including prompt refinement, attention re-weighting, and word swapping, similar to the capabilities of image-P2P. Video-P2P is able to maintain semantic consistency across different frames and preserve the temporal coherence of the original video during the editing process. More examples can be found in the appendix.

\paragraph{Word swap.}
Video-P2P enables the replacement of entities based on word swapping while maintaining the coherence of unrelated regions. As illustrated in Fig.~\ref{fig:examples}, Video-P2P seamlessly replaces the man on the motorbike with Spider-Man while minimizing the changes to the motorbike's appearance (the 4th row). The generated Spider-Man exhibits a consistent appearance across frames, and the background remains unchanged. Furthermore, we can replace a dog with a cat while preserving its gesture and the surrounding grass (the 5th row).

\paragraph{Prompt refinement.}
Video-P2P is able to do prompt refinement, such as modifying object properties.
For example, we can transform the running dog into a robotic one (the 6th row in Fig.~\ref{fig:examples}), and convert a motorbike into a Lego toy with the same motion (the 3rd row).
Notice the grass and sky are almost not influenced.
Additionally, Video-P2P can perform global editing like changing the weather to sunset or flooding the road with water (2nd row). Style transfer can also be accomplished by Video-P2P, as exemplified by transforming the video into a watercolor painting.

\paragraph{Attention re–weighting.}
Similar to Image-P2P, Video-P2P also enables attention re-weighting. By adjusting the cross-attention of specific words, we can manipulate the extent of the corresponding generation.
For instance, we can regulate how fluffy a dog is in the video (the 6th row of Fig.~\ref{fig:examples}).

\subsection{Comparison}
\paragraph{Comparison with Tune-A-Video.}
Both TAV+DDIM~\cite{wu2022tune} and our Video-P2P allow for video editing with text prompts. However, TAV+DDIM cannot avoid altering the entire video content when editing specific objects, while Video-P2P can edit a local area and minimize the influence on other regions. Fig.~\ref{fig:p2p_vs_tav} (Left) demonstrates that Video-P2P preserves the complex shape of the cloud when replacing a lion with King Kong, whereas TAV+DDIM can only maintain the color tone of the sky in this case.

Although our model initialization is similar to TAV, Video-P2P can still generate temporal-consistent results where TAV+DDIM fails. As demonstrated in Fig.~\ref{fig:p2p_vs_tav} (Right), TAV struggles to generate a temporally consistent sequence in the second row, even when the inputs are features from DDIM inversion. In contrast, our method can produce better structure-preserved results, as shown in the third row.

\begin{figure}
\begin{center}
    \includegraphics[width=1\linewidth]{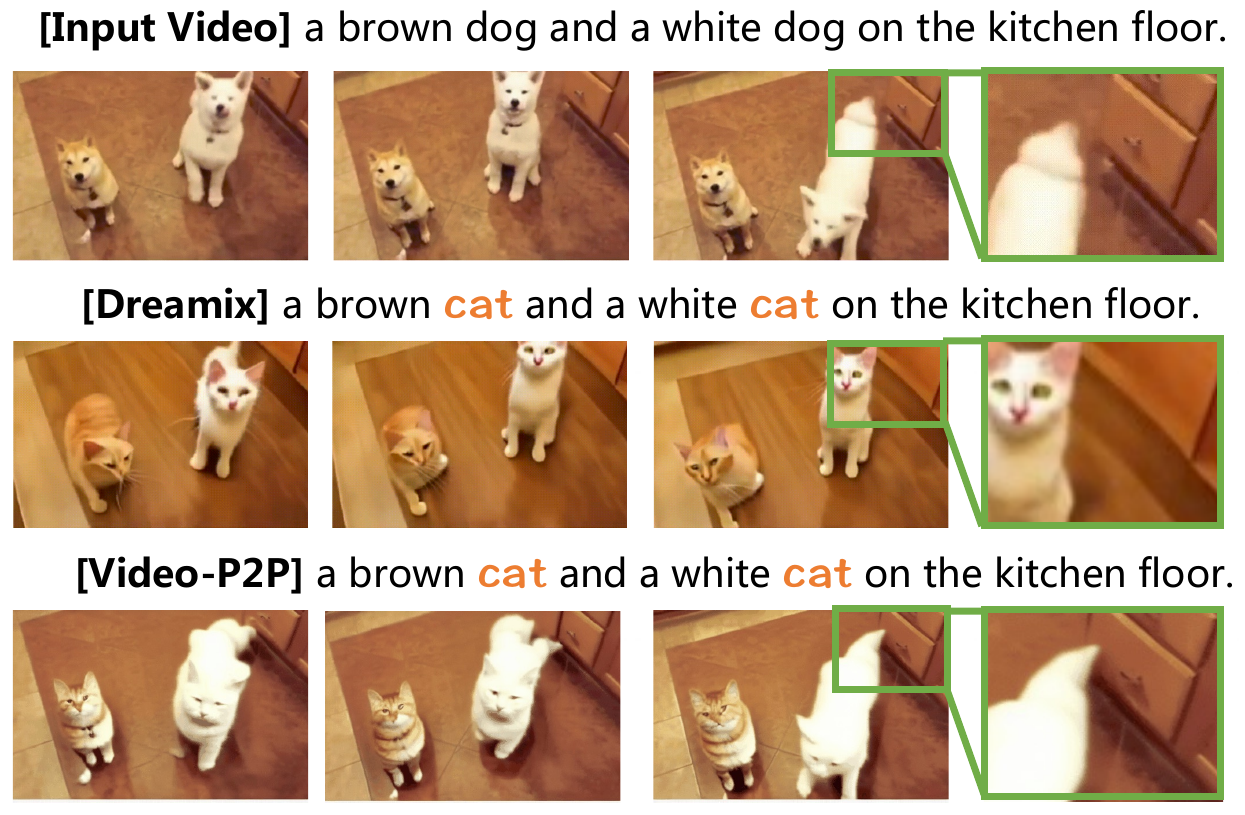}
\end{center}
\vspace{-14pt}
\caption{Video-P2P v.s. Dreamix. Both methods can change the objects to new categories. Only Video-P2P can preserve the details in the background.}
\label{fig:p2p_vs_dreamix}
\vspace{-6pt}
\end{figure}

\paragraph{Comparison with Dreamix.}
In contrast to Dreamix~\cite{molad2023dreamix}, which uses a pre-trained video diffusion model that is not publicly available, our method yields superior results for subject replacement. Although our method cannot perform video motion editing due to the lack of temporal priors, we outperform Dreamix in preserving details and motion consistency. As Dreamix is not open-sourced, we conducted our evaluation on its released demo. As demonstrated in Fig.~\ref{fig:p2p_vs_dreamix}, both methods can transform two dogs into two cats, but our method preserves the details of the drawer in the background (the 3rd row). Furthermore, Dreamix may affect the time sequence to some extent, as the generated cat moves more slowly than the original dog in the video. In contrast, our method completely preserves the motion of the original video.

\paragraph{Quantitative results.}
We evaluate our proposed Video-P2P on 10 YouTube videos and report four metrics for quantitative analysis. The CLIP Score measures the textual similarity between the text prompt and video, while Masked PSNR and LPIPS~\cite{zhang2018lpips} evaluate the quality of structure preservation. We also proposed a novel metric, Object Semantic Variance (OSV), to measure semantic consistency across frames. For detailed explanations of these metrics, please refer to the appendix. Our results, as shown in Table~\ref{tab:quant_compare}, demonstrate that Video-P2P performs well on all metrics. Compared to TAV+DDIM, Video-P2P achieves higher Masked PSNR and lower LPIPS, indicating better preservation of unchanged regions. Compared to the other two methods, Video-P2P has a much lower OSV, indicating its superior ability to maintain semantic consistency across frames.
Moreover, in Tab.~\ref{tab:user_study}, we report the user study results, where Video-P2P ranks first on average and has a high preference rate compared to other methods.

\begin{figure}
\begin{center}
   \includegraphics[width=1\linewidth]{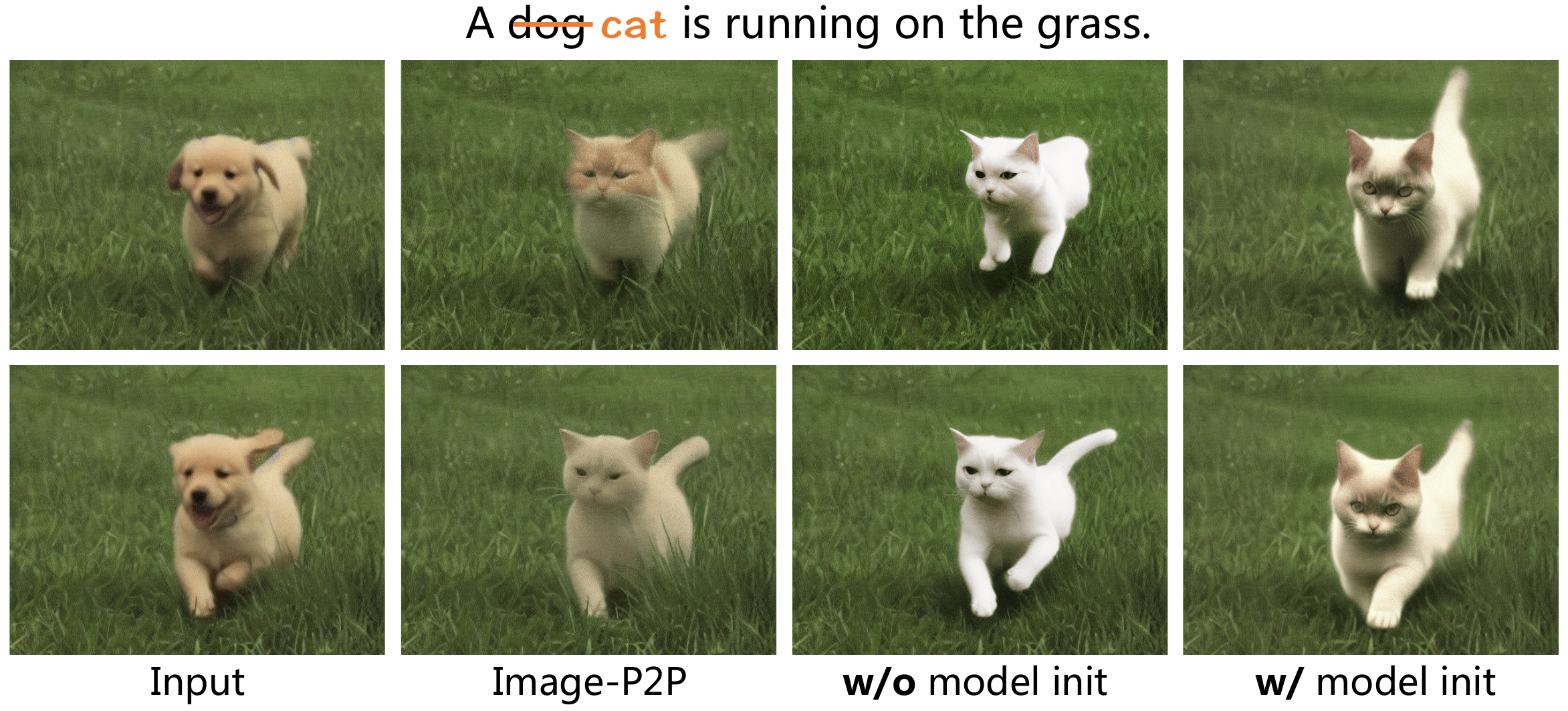}
\end{center}
\vspace{-13pt}
   \caption{Ablation on model initialization. Though Video-P2P can still generate consistent content without model initialization, the generation quality is degraded. With the model initialization, the generation quality is recovered.}
\label{fig:init_ablation}
\vspace{-7pt}
\end{figure}

\begin{figure}
\begin{center}
   \includegraphics[width=1\linewidth]{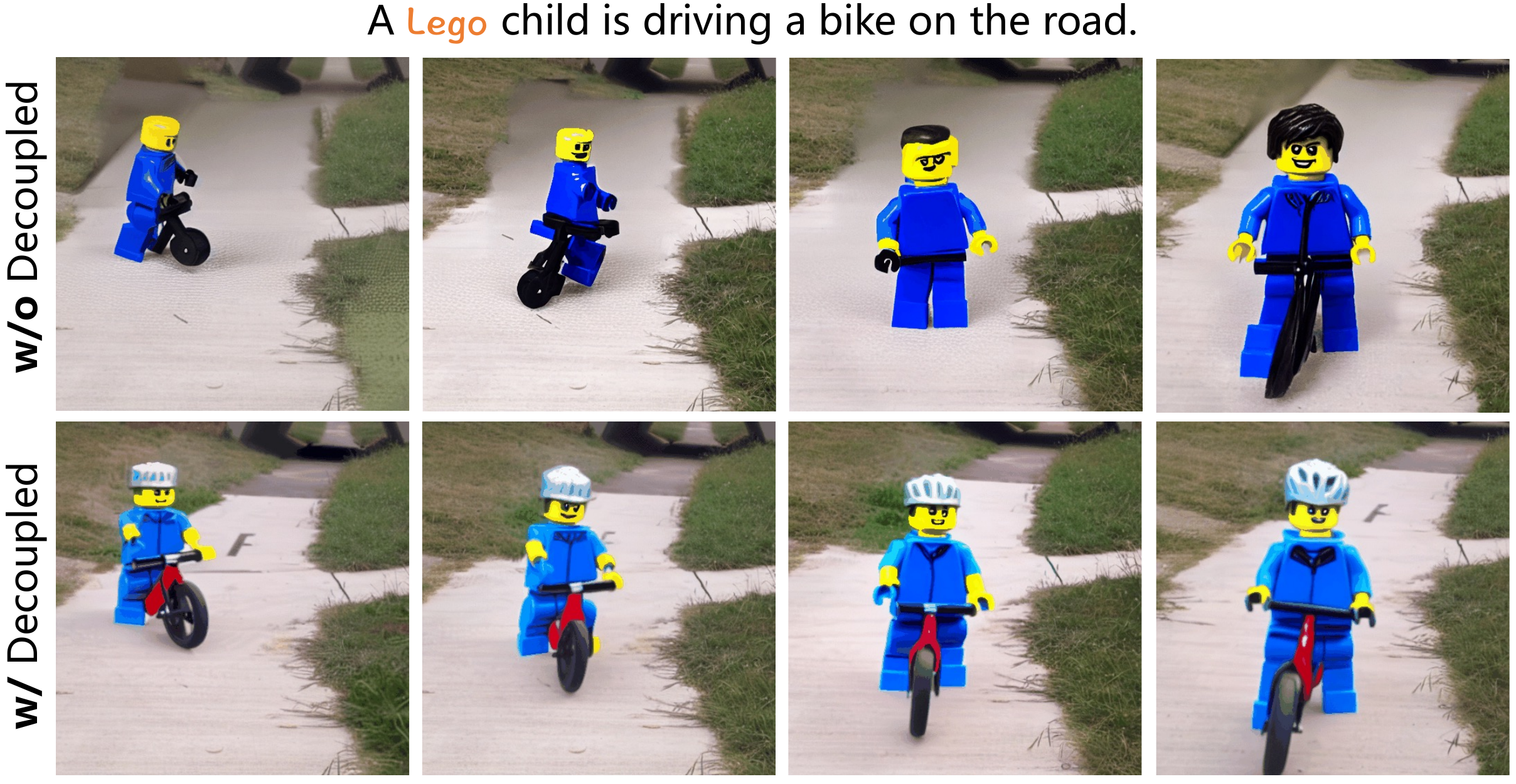}
\end{center}
\vspace{-10pt}
   \caption{Ablation on decoupled-guidance attention control. Using decoupled-guidance improves the generation quality.}
\label{fig:inversion_ablation}
\vspace{-5pt}
\end{figure}

\subsection{Ablation Study}
\vspace{-2mm}
\paragraph{Model initialization.}
While the inflated image diffusion model can generate semantically consistent images, the T2S model's generation ability is compromised during inflation, making it inadequate for video inversion even with an optimized unconditional embedding. As seen in Fig.\ref{fig:init_ablation} (the 3rd column), directly using the inflated T2S model produces unrealistic results with an inaccurate background. To mitigate this, we initialize the T2S mode by fine-tuning the given video. This is evident in Fig.\ref{fig:init_ablation} (4th column), where the cat's appearance improves, and the grass reconstruction becomes more accurate.

\paragraph{Shared unconditional embedding.}
Table~\ref{tab:inversion_compare} presents the quantitative results for video inversion. We observe that optimizing a shared unconditional embedding can significantly improve the PSNR compared to TAV+DDIM. However, using multiple unconditional embeddings for each frame only increases the PSNR by 0.2 but results in a higher parameters usage ($n$ times). Besides, we find that using multiple unconditional embeddings leads to a lower Masked PSNR of 20.51 after attention control compared to the shared unconditional embedding. Thus, we conclude that shared unconditional embedding is the most effective and efficient method for video inversion.

\paragraph{Decoupled-guidance attention control.}
To obtain the latent features of the input video, we optimize an unconditional embedding using the source prompt. It is important to note that this embedding is only suitable for the source prompt during the prompt-to-prompt process. Using the optimized embedding for the target prompt may negatively impact the quality of the generated results, as shown in Fig.~\ref{fig:inversion_ablation} (1st row).
Instead, we utilize the initialized unconditional embedding for the target prompt and incorporate attention maps from two branches. The decoupled-guidance attention control approach significantly improves the editing quality, as shown in Fig.\ref{fig:inversion_ablation} (the 2nd row). Quantitative ablations can be found in Tab.~\ref{tab:quant_compare} (the 3rd row and 4th row).

\begin{table}
\footnotesize
\begin{center}
\renewcommand\arraystretch{1.1}
	{
		\begin{tabular}{y{50}x{33}x{33}x{33}x{31}}
            \hline
            & CLIP$~{\textcolor{purple}{\uparrow}}$ &  M.PSNR$~{\textcolor{purple}{\uparrow}}$ & LPIPS$~{\textcolor{purple}{\downarrow}}$  & OSV$~{\textcolor{purple}{\downarrow}}$ \\
            \hline
            TAV+DDIM & 0.3322 & 17.32 & 0.4625 & 55.12\\
            Image-P2P & 0.3269 & 22.99 & 0.3065 & 76.50 \\
            Ours (w/o DG) & 0.3224 & 18.96 & 0.3864 & 68.76 \\
            \textbf{Ours}  & \cellcolor{Gray}0.3361 & \cellcolor{Gray}20.54 & \cellcolor{Gray}0.3297 & \cellcolor{Gray}47.57 \\
            \hline
            \end{tabular}
        }
\end{center}
\vspace{-10pt}
\caption{Quantitative evaluation. We evaluate editing textual similarity (CLIP Score), region preservation (Masked PSNR, LPIPS), and Object Semantic Variance~(OSV) for semantic consistency. DG refers to Decoupled-Guidance.}
\label{tab:quant_compare}
\end{table}

\begin{table}
\footnotesize
\begin{center}
\renewcommand\arraystretch{1.1}
	{
		\begin{tabular}{y{33}m{35pt}<{\centering}m{35pt}<{\centering}m{35pt}<{\centering}m{42pt}<{\centering}}
            \hline
            & VQVAE & TAV +DDIM & Multi-uncond  & \textbf{Shared-uncond} \\
            
            \hline
            PSNR(dB)$~{\textcolor{purple}{\uparrow}}$ & 24.73 & 15.43 & 22.97 & \cellcolor{Gray}22.75\\
            \#Param.$~{\textcolor{purple}{\downarrow}}$ & / & 0.13M & 22.68M & \cellcolor{Gray}2.94M \\
            \hline
            \end{tabular}
        }
\end{center}
\vspace{-10pt}
\caption{Reconstruction quality on video inversion. A shared unconditional embedding can reconstruct a high-quality video with a small size.}
\vspace{-3pt}
\label{tab:inversion_compare}
\end{table}

\begin{table}
\footnotesize
\begin{center}
\renewcommand\arraystretch{1.1}
	{
		\begin{tabular}{y{40} cccc}
            \hline
            & Image-P2P & TAV & TAV+DDIM  & \textbf{Video-P2P} \\
            \hline
            Structure   & \underline{2.67}      & \underline{3.33}    & \underline{2.61}    & \cellcolor{Gray}\underline{1.39}\\
            Preserving  & 13.59\%   & 6.52\%  & 10.87\% & \cellcolor{Gray} 69.02\% \\
            \arrayrulecolor{DarkGray}\hline
            Text        & \underline{3.40}      & \underline{2.78}    & \underline{2.28}    & \cellcolor{Gray}\underline{1.54}\\
            Alignment   & 3.80\%   & 14.13\%  & 19.57\%  & \cellcolor{Gray} 62.50\% \\
            \arrayrulecolor{DarkGray}\hline
            Realism \&  & \underline{3.38}      & \underline{2.98}   & \underline{2.21}    & \cellcolor{Gray}\underline{1.43}\\
            Quality  &  4.35\%   & 7.61\% & 19.02\% & \cellcolor{Gray} 69.02\% \\
            \arrayrulecolor{black}\hline
            \end{tabular}
        }
\end{center}
\vspace{-10pt}
\caption{User study result of \underline{average ranking}$~{\textcolor{purple}{\downarrow}}$ and preference rate$~{\textcolor{purple}{\uparrow}}$.}
\label{tab:user_study}
\vspace{-3pt}
\end{table}

\section{Conclusion}
Our proposed approach, Video-P2P, provides a simple yet effective solution for video editing with cross-attention control. By leveraging a pre-trained image diffusion model, we demonstrate that editing a video locally and globally is possible. Specifically, we optimize a shared unconditional embedding based on a well-initialized T2S model for video inversion. We also propose using different unconditional embeddings for source and target prompts, and integrating attention maps from two branches for improved attention control. These techniques enable Video-P2P to perform various applications, such as word swap, prompt refinement, and attention re-weighting. In future work, we will enhance its capability to handle more complex editing tasks like injecting extra objects.

{\small
\bibliographystyle{ieee_fullname}
\bibliography{egbib}
}

\end{document}